\documentclass{article}

\usepackage[preprint]{neurips_2026}

\usepackage[utf8]{inputenc} 
\usepackage[T1]{fontenc}    
\usepackage{hyperref}       
\usepackage{url}            
\usepackage{booktabs}       
\usepackage{amsfonts}       
\usepackage{nicefrac}       
\usepackage{microtype}      
\usepackage{xcolor}         
\usepackage{amsmath}
\usepackage{enumitem}

\usepackage{pifont}

\usepackage{colortbl}
\usepackage{bm}     
\usepackage{xfrac} 
\usepackage{array}
\usepackage{makecell}
\usepackage{multirow}
\usepackage{wrapfig}
\usepackage{bbding}
\usepackage{cleveref}
\usepackage{adjustbox}

\usepackage{booktabs}
\usepackage[table]{xcolor}
\usepackage{pifont}
\newcommand{\cmark}{\ding{51}}
\newcommand{\xmark}{\ding{55}}

\usepackage{floatrow}
\newfloatcommand{capbtabbox}{table}[][\FBwidth]
\usepackage{wrapfig}

\definecolor{mygray}{gray}{.9}
\definecolor{myyellow}{rgb}{0.71, 0.55, 0.0}
\definecolor{mygreen}{rgb}{0.0, 0.51, 0.0}
\definecolor{myblue}{rgb}{0.0, 0.71, 0.71}
\definecolor{color2}{rgb}{0.55, 0.71, 0.0}
\definecolor{color1}{rgb}{0.98, 0.81, 0.69}
\definecolor{color3}{rgb}{1.0, 0.6, 0.4}
\definecolor{color4}{rgb}{0.29, 0.59, 0.82}
\title{Bridging Domain Gaps in Few-Shot Object Detection via Prompt-Guided Generation and Feature Perturbation}
\title{Prompt-Driven Domain Simulation with Feature Perturbation for Cross-Domain Few-Shot Object Detection}
\title{Prompt-Driven Simulation with Feature Perturbation for Cross-Domain Few-Shot Object Detection}

\author{%
  Linhai Zhuo \\
  Fuzhou University\\
  \texttt{linhaizhuo@fzu.com}
  \And
  Junxi Cai\thanks{Co-author.}\\
  Fuzhou University\\
  \texttt{2501027223@fzu.edu.cn}
  \And
  Tianwen Qian\\
  East China Normal University\\
  \texttt{twqian@cs.ecnu.edu.cn}
  \And
  Qingping Zheng\thanks{Corresponding author.}\\
  Xiamen University\\
  \texttt{zhengqingping@xmu.edu.cn}
  \And
  Yang Liu\\
  King's College London\\
  \texttt{yang.15.liu@kcl.ac.uk}
}

\begin{document}

\maketitle

\begin{abstract}
Data augmentation, which simulates diverse visual variations to expand the source distribution and induce synthetic domain shifts, is a simple yet effective strategy for mitigating severe domain shifts and limited labeled target data in cross-domain few-shot object detection (CD-FSOD).
Existing approaches rely on conventional data augmentation, such as Color-Jitter, Mosaic, and background-centric adaptation (e.g., Domain-RAG), which are limited in modeling complex domain shifts and often lead to suboptimal performance.
In this paper, we propose PSP-FSOD, a principled framework that integrates prompt-driven domain simulation with feature perturbation regularization to improve generalization in CD-FSOD.
To enable controllable domain synthesis, we design a prompt-driven strategy that leverages the visual grounding capability of large VLMs to jointly model foreground and background variations, generating semantically consistent yet domain-diverse training samples. Moreover, we adopt a grounding-aware generation scheme that guides object placement and alleviates semantic–spatial misalignment, thereby improving foreground adaptation.
To ensure training stability and robustness, we further introduce a noise-induced feature perturbation mechanism that injects Gaussian noise into multi-scale intermediate features with distribution correction, encouraging consistent predictions under perturbations and reducing reliance on domain-specific cues.
Extensive experiments demonstrate that PSP-FSOD produces high-quality domain-diverse supervision and learns domain-invariant representations, consistently improving performance across CD-FSOD benchmarks.
\end{abstract}
\section{Introduction}

Cross-domain few-shot object detection (CD-FSOD)~\cite{fu2024cross} aims to adapt object detectors to novel target domains with only a handful of annotated examples, reflecting a realistic yet challenging setting in practical applications. This problem is inherently difficult due to the simultaneous presence of data scarcity and domain shift. In particular, a detector trained on source domains with abundant labeled data may fail to generalize to target domains with significantly different distributions, such as artistic styles like clipart or sketch, despite sharing the same semantic categories. As a result, the model is required to learn feature representations that are both domain-invariant and class-discriminative under extremely limited supervision in the target domain. To mitigate domain shift, data augmentation simulates diverse visual variations (e.g., context, texture, and style changes), improving robustness by expanding the source distribution via input- and feature-level perturbations and exposing the model to synthetic domain shifts.

An intuitive data augmentation strategy is Copy-Paste~\cite{zhao2023x} (Fig. 1(a)), which increases training diversity by compositing object instances across images. However, such naive composition often breaks contextual realism, yielding weak domain coherence and poor approximation of the target distribution. As a result, it tends to introduce synthetic artifacts rather than effectively reducing the source–target gap.
In parallel, conventional augmentations such as Color-Jitter (Fig. 1(b)) and Mosaic, along with search-based methods like ETS~\cite{pan2025enhance} (Fig. 1(c)), enhance data diversity through predefined transformations. Yet their limited augmentation space restricts their ability to capture complex domain variations, resulting in suboptimal generalization under domain shift.
More recently, modern generative models (e.g., SDXL~\cite{podell2023sdxl} and FLUX~\cite{flux2024}) have demonstrated remarkable capability in synthesizing high-fidelity images. Directly leveraging these models remains non-trivial, as unconstrained generation can disrupt object grounding, leading to misalignment between semantic categories and their corresponding spatial regions.
To mitigate this issue, Domain-RAG~\cite{li2025domainrag} builds upon generative models to improve domain diversity through background-level adaptation while preserving foreground objects (Fig. 1(d)). Nevertheless, this background-centric strategy limits foreground adaptation by keeping objects unchanged and ignoring variations in viewpoint, pose, and appearance. Consequently, the fixed foreground representation struggles to generalize across diverse object instances (e.g., intra-class variations such as bees and hornets), thereby limiting its effectiveness for CD-FSOD.
\begin{figure}
    \centering
    \includegraphics[width=\linewidth]{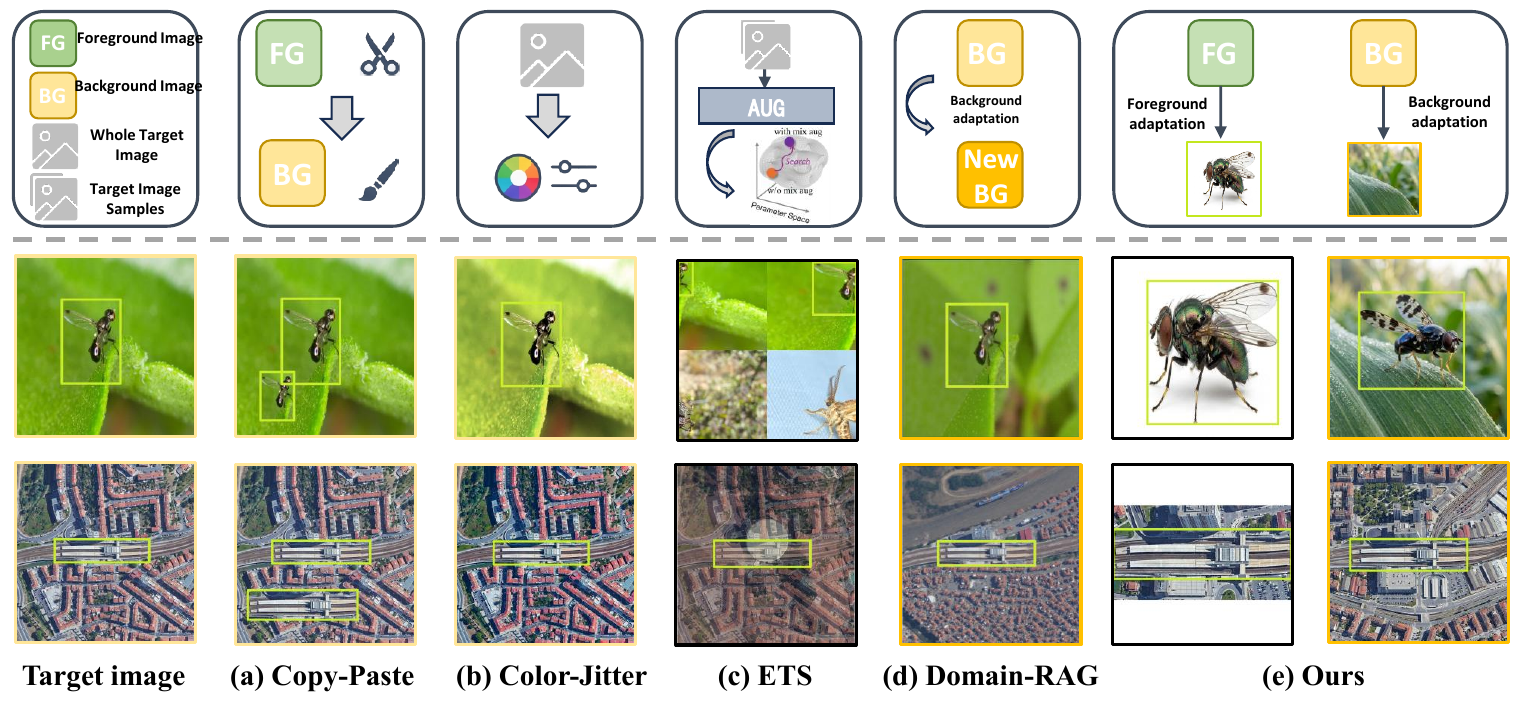}
    \caption{Comparison of domain data simulation strategies: (a) Copy-Paste, which introduces limited diversity and often breaks contextual consistency; (b) Color-Jitter, which applies low-level appearance changes but fails to model complex domain shifts; (c) ETS, which searches within predefined transformation spaces and lacks semantic controllability; (d) Domain-RAG, which adapts backgrounds while keeping foregrounds unchanged, limiting foreground–background interaction; and (e) our prompt-driven domain simulation, which jointly adapts foreground and background to generate semantically consistent yet domain-diverse samples for cross-domain adaptation.}
    \label{fig:placeholder}
\end{figure}

Inspired by the strong visual grounding of Vision-Language Models (VLMs), such as Qwen~\cite{yang2025qwen3}, Gemini~\cite{team2023gemini}, and GPT-4~\cite{achiam2023gpt}, which enable fine-grained alignment between language and visual regions, they offer a promising paradigm for modeling foreground variations and background contextual diversity.
In this paper, we propose a \textbf{\textit{prompt-driven domain simulation strategy}} that leverages VLMs as a data augmentation engine to model domain variations. 
Concretely, we design a semantic consistency-constrained prompt to guide Gemini in preserving foreground semantics (e.g., object category and structure) while varying background attributes such as context, style, and texture. This enables controllable synthesis of domain-diverse samples with semantic consistency. 
Furthermore, for foreground variation, we first generate a visually distinct bounding box to indicate the target object and integrate it with the background to control spatial placement. This better leverages the grounding capability of VLMs, preserving spatial correspondence and mitigating semantic–spatial misalignment in direct image generation. 
This yields high-quality, domain-diverse supervision, improving robustness and generalization for cross-domain few-shot object detection.
To ensure training stability, we introduce \textbf{\textit{noise-induced feature perturbation}} as a complementary regularization. We inject Gaussian noise into intermediate multi-scale features with a distribution correction step, encouraging stable predictions under perturbations and reducing sensitivity to domain-specific cues. This promotes learning domain-invariant and class-discriminative representations, which is particularly important in the few-shot setting.
Overall, we propose a unified framework, \textbf{P}rompt-Driven \textbf{S}imulation with Feature \textbf{P}erturbation for Cross-Domain \textbf{F}ew-\textbf{S}hot \textbf{O}bject \textbf{D}etection (PSP-FSOD), which combines prompt-driven domain simulation and noise-induced feature perturbation to improve data diversity, representation robustness, and generalization. Our contributions are summarized as follows:
\begin{itemize}[leftmargin=20pt]
    \item We introduce a prompt-driven domain simulation strategy that leverages the visual grounding capability of VLMs to enable coordinated foreground–background adaptation, generating semantically consistent yet domain-diverse samples for effective cross-domain few-shot learning.

    \item We develop a feature perturbation mechanism as a regularization strategy that injects Gaussian noise into multi-level feature representations, followed by a distribution correction step, to improve robustness to domain shifts and promote domain-invariant representation learning.

    \item We further propose PSP-FSOD, a unified framework that integrates these components to improve semantic consistency and feature robustness, achieving state-of-the-art results on benchmarks. Built on VLM-based domain simulation, it consistently improves results across different shots (1-shot: 34.2→36.4, 5-shot: 44.3→45.0, 10-shot: 46.6→47.1).
\end{itemize}

\section{Related Works}
\paragraph{Cross-domain Few-shot Object Detection.}
Cross-domain few-shot object detection (CD-FSOD) aims to adapt detectors to target domains with only a few annotated samples under severe domain shifts. Early studies explore feature alignment, domain-adaptive detection, data augmentation, and knowledge distillation, including FAFR-CNN~\cite{wang2019few}, PICA~\cite{zhong2022pica}, DAPN~\cite{zhao2021domain}, AcroFOD~\cite{gao2022acrofod}, and Distill-CDFSOD~\cite{xiong2023cd}. Recent methods can be broadly categorized into restricted-source and open-resource settings. The former, e.g., CD-ViTO~\cite{fu2024cross}, typically constrains source training data to fixed datasets such as MS-COCO~\cite{lin2014microsoft}. The latter leverages foundation models for data augmentation or generalization enhancement. For instance, ETS~\cite{pan2025enhance} searches effective augmentations with GroundingDINO~\cite{liu2024grounding}, while Domain-RAG~\cite{li2025domainrag} synthesizes target-style samples by adapting backgrounds. Other approaches improve fine-tuning strategies, feature decorrelation, or attention regularization~\cite{meng2025cdformer}.
Unlike prior generation-based methods, our approach improves semantic generalization via prompt-driven image synthesis and enhances fine-grained robustness via noise-induced feature perturbation.

\paragraph{Data Augmentation.}
Data augmentation is a key strategy for mitigating annotation scarcity and domain shifts in CD-FSOD. Conventional approaches rely on image-level transformations such as flipping, cropping, resizing, color jittering, Mosaic, Copy-Paste, and MixUp~\cite{zhang2017mixup, fu2021meta}. However, these simple operations are often insufficient to model realistic target-domain variations while preserving both semantic consistency and domain style.
Recent works have explored more effective augmentation strategies for CD-FSOD. ETS~\cite{pan2025enhance} leverages foundation models such as GroundingDINO~\cite{liu2024grounding} to search for effective augmentation configurations.
Domain-RAG~\cite{li2025domainrag} instead adopts a generative augmentation paradigm, using retrieval-guided compositional image generation to synthesize target-style backgrounds while preserving foreground semantics.
Unlike these methods, our approach leverages prompt design and the visual grounding capability of VLMs to enable coordinated foreground–background adaptation, generating semantically consistent yet domain-diverse samples.

\paragraph{Noise Enhanced Generalization.}
Noise perturbation has been widely adopted to improve robustness and generalization under distribution shifts~\cite{taniguchi2024learning, bar2024frozen}. In domain generalization, perturbations are applied at either the image/style level or feature-statistics level to simulate unseen domains, including feature style randomization~\cite{wang2022feature}, normalization-guided augmentation~\cite{qi2024normaug}, and adversarial style augmentation~\cite{fu2022wave, fu2023styleadv}. In test-time adaptation, feature perturbations are further used to exploit unlabeled target data via consistency regularization~\cite{cho2025feature}. Related ideas have also been explored in cross-domain few-shot learning and object detection, where input noise, adversarial feature perturbations, patch-wise Gaussian noise, and frequency-domain noise improve robustness to domain shifts and corruptions~\cite{liang2021boosting,ma2020training,lopes2019improving,vaish2024fourier}.
In contrast, we couple noise perturbation with the generation process and further inject noise into multi-scale features with a subsequent correction step, improving robustness to fine-grained details.

\begin{figure}
    \centering
    \includegraphics[width=\linewidth]{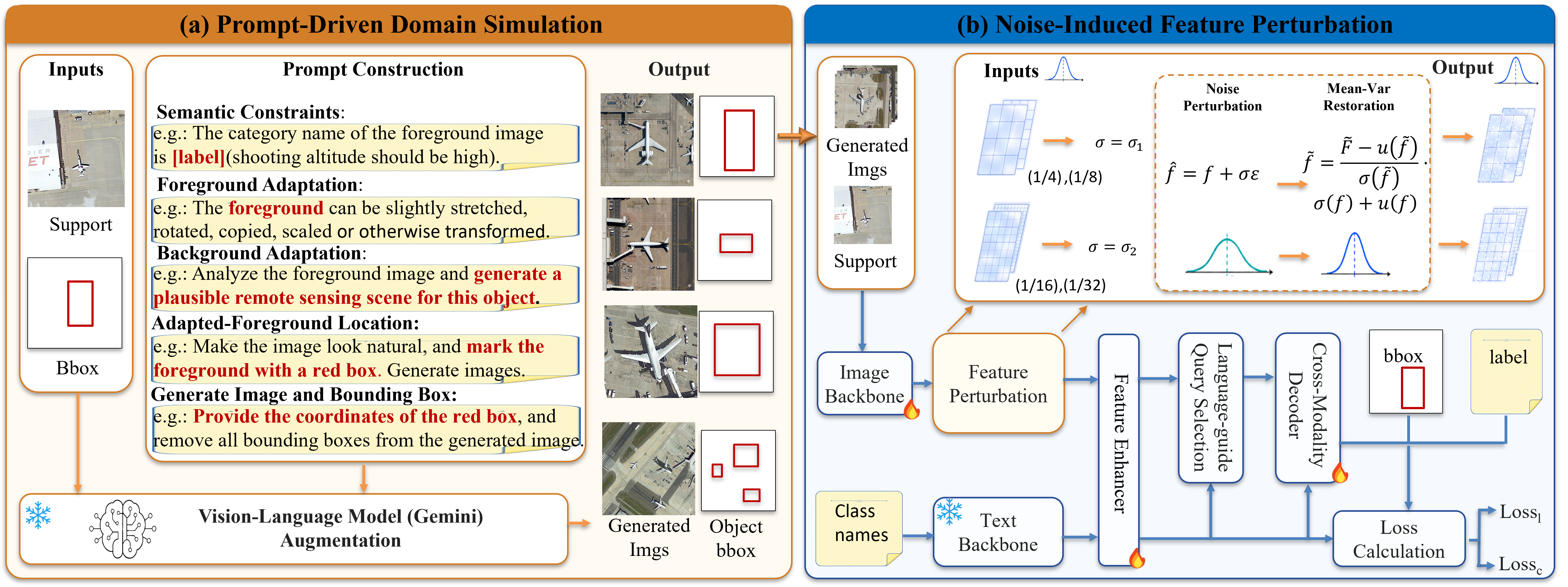}
    \caption{
    Overview of the proposed framework. 
    (a) Prompt-Driven Domain Simulation generates diverse and plausible target-domain images using structured prompts, support image and bbox. 
    (b) Noise-Induced Feature Perturbation injects feature-level noise into GroundingDINO and applies restoration to improve robustness.
    }
    \label{fig_framework}
\end{figure}

\section{Method}
Cross-domain few-shot object detection aims to adapt a pretrained detector to a target domain under extremely limited labeled data.
Given a target domain $\mathcal{D}_{T}$, under the standard $N$-way $K$-shot protocol (with $N$ target categories), the support set $\mathcal{S}^{N \times K} \subset \mathcal{D}_{T}$ contains $K$ annotated instances per category, and each image is associated with bounding-box annotations and category labels. 
A disjoint query set $\mathcal{Q}$ is used for evaluation.
Since no additional real training data are available, existing approaches for addressing cross-domain shift predominantly rely on data augmentation strategies. These include traditional appearance-level transformations~\cite{pan2025enhance} such as color jittering, geometric augmentations, and copy-paste, as well as background-centric methods such as Domain-RAG~\cite{li2025domainrag}. However, these methods remain suboptimal, as they mainly introduce low-level visual variations and struggle to model complex cross-domain distribution shifts. 

To overcome this limitation, we leverage a vision-language model (VLM) to synthesize auxiliary training samples under semantic consistency constraints, and fine-tune the detector on the union of real support data $\mathcal{S}^{N \times K}$ and generated samples $\mathcal{S}_{G}$, enabling more effective adaptation to the target domain $\mathcal{D}_{T}$. To further enhance robustness to fine-grained target-domain variations, we introduce noise-induced feature perturbation during training.
Consequently, we propose a unified framework, PSP-FSOD, for cross-domain few-shot object detection by jointly exploiting image-level domain simulation and feature-level robustness learning.
The overall pipeline is illustrated in Fig. \ref{fig_framework}, which consists of two key components: Prompt-Driven Domain Simulation under semantic consistency constraints and Noise-Induced Feature Perturbation for learning domain-invariant representations. The details of each component are provided below.



\subsection{Prompt-Driven Domain Simulation under Semantic Consistency Constraints}

To improve semantic generalization in cross-domain few-shot object detection, we leverage Gemini~\cite{team2023gemini} as a training-free vision-language augmentation engine to simulate target-domain variations from the limited support set.
As shown in Fig.~\ref{fig_framework}(a), our prompt-driven domain simulation (PDS) synthesizes images by leveraging a vision-language model to explicitly control semantic variations in both foreground and background while preserving overall semantic consistency.
Specifically, each support image and its bounding-box annotation are used to extract the foreground object while preserving category semantics. A structured prompt then guides the model to maintain object identity and structural integrity, while enabling controllable variations in appearance, background context, and spatial layout. In addition, we exploit the visual grounding capability of the VLM to ensure alignment between the generated foreground and surrounding regions, thereby reducing foreground-background mismatch and improving spatial coherence in the synthesized samples.

Given a target-domain support image \(I\) with its bounding-box annotation \(B\), we use \(B\) to localize the foreground object, which serves as a semantic anchor during generation. The category label \(y\) and the target-domain description \(d\) are encoded into a structured prompt \(P(y,d)\), which is fed into the Gemini VLM to generate augmented images and annotations. The process is formulated as:
\[
(\hat{I}, \hat{B})=\mathcal{G}\big(I,B;P(y,d)\big),
\]
where \(\hat{I}\) and \(\hat{B}\) denote the generated image and its bounding-box annotation, respectively. The target-domain description \(d\) captures domain-specific characteristics and can be obtained by a large language model through summarizing dataset papers, benchmark descriptions, and a small set of target-domain samples. It typically includes imaging conditions, visual styles, scene contexts, texture properties, and spatial relationships.
The structured prompt is decomposed into five parts:
\[
P(y,d)=
\{
P_{\mathrm{sem}},
P_{\mathrm{fg}},
P_{\mathrm{bg}},
P_{\mathrm{loc}},
P_{\mathrm{ann}}
\},
\]
where \(P_{\mathrm{sem}}\), \(P_{\mathrm{fg}}\), \(P_{\mathrm{bg}}\), \(P_{\mathrm{loc}}\), and \(P_{\mathrm{ann}}\) correspond to Semantic Constraints, Foreground Adaptation, Background Adaptation, Adapted-Foreground Location, and Generate Image and Bounding Box in Fig.~\ref{fig_framework}(a), respectively. 
Specifically, \ding{172} \textbf{Semantic Constraints} preserves the semantic consistency of the generated object by explicitly specifying its category label. 
\ding{173} \textbf{Foreground Adaptation} allows controlled changes to the foreground while keeping its category semantics unchanged. 
\ding{174} \textbf{Background Adaptation} guides the model to generate a target-domain scene compatible with the given object. 
\ding{175} \textbf{Adapted-Foreground Location} constrains the placement of the adapted foreground in the generated scene, so that the image remains natural and spatially reasonable.
\ding{176} \textbf{Generate Image and Bounding Box} further requires the model to return usable object annotations. 
In practice, the visible red box is only used as an intermediate step to obtain the object coordinates, and the final generated image is cleaned by removing all visual boxes.
These modules jointly regulate semantic preservation, foreground and background variations, object placement, and annotation generation during the synthesis process.

Finally, each generated sample is represented as \((\hat{I},y,\hat{B})\) and added to the target-domain support set for subsequent detector fine-tuning. Through this PDS process, the proposed method generates category-consistent, scene-adapted, and spatially plausible augmented samples, providing richer and more reliable target-domain supervision for cross-domain few-shot object detection.

\subsection{Noise-Induced Feature Perturbation for Domain-Invariant Representations}

To promote domain-invariant representations under fine-grained target-domain variations, we introduce Noise-Induced Feature Perturbation (NFP) at the visual feature level. Specifically, we inject noise into the extracted features and regularize the perturbed representations by preserving their mean and variance, thereby maintaining consistent feature statistics across domains. As shown in Fig.~\ref{fig_framework}(b), NFP is inserted between the backbone output and the feature enhancer in GroundingDINO~\cite{liu2024grounding}, i.e., applied after multi-level feature extraction and before cross-scale feature enhancement.

Technically, during training, the input image is first encoded by the GroundingDINO image backbone (Swin Transformer~\cite{liu2021swin}) into a four-level feature pyramid, denoted as \(\{X_i^{\mathrm{clean}}\}_{i=1}^{4}\). 
Lower-level features capture local details such as edges, textures, and object boundaries, while higher-level features encode more abstract semantics related to object categories and global structure. Motivated by this hierarchical property, we inject noise at different strengths across feature levels to account for their varying sensitivity to perturbations.
Before being fed into the feature enhancer, Gaussian noise with the same shape as each feature map is sampled and added to the corresponding clean feature. The resulting noisy feature at the \(i\)-th level is formulated as:
\[
X_i^{\mathrm{noisy}} = X_i^{\mathrm{clean}} + \sigma_i \epsilon_i,
\quad
\epsilon_i \sim \mathcal{N}(0,I),
\]
where \(\sigma_i\) denotes the noise strength for the \(i\)-th feature level. Notably, the injection probability is set to 1, meaning that visual features are perturbed in every forward pass during training.



Following the hierarchical structure of the feature pyramid, we divide the four feature levels into two groups and assign different noise strengths:
\[
\sigma_i =
\begin{cases}
\sigma_1, & i \in \{1,2\},\\
\sigma_2, & i \in \{3,4\},
\end{cases}
\quad
\sigma_1=0.15,\quad \sigma_2=0.10.
\]
Specifically, the lower-level features (resolutions $1/4$ and $1/8$) are assigned a larger noise strength $\sigma_1$ to improve robustness to low-level appearance variations. In contrast, the higher-level features (resolutions $1/16$ and $1/32$) use a smaller noise strength $\sigma_2$ to preserve high-level semantic representations and avoid excessive feature corruption.

Since directly adding Gaussian noise may shift the original feature distribution and destabilize training, we further introduce a distribution correction step before feeding the perturbed features into the feature enhancer. 
Concretely, the noisy feature is first normalized using its own statistics and then re-scaled using the mean and variance of the corresponding clean feature:
\[
X_i^{\mathrm{noisy}} \leftarrow
\operatorname{cast}_{\operatorname{dtype}(X_i^{\mathrm{clean}})}
\left(
\frac{
X_i^{\mathrm{noisy}} - \mu(X_i^{\mathrm{noisy}})
}{
\sigma(X_i^{\mathrm{noisy}})
}
\cdot \sigma(X_i^{\mathrm{clean}})
+ \mu(X_i^{\mathrm{clean}})
\right).
\]
Here, \(\mu(\cdot)\) and \(\sigma(\cdot)\) denote the mean and standard deviation of a feature map, respectively.
Overall, this design enables noise injection to introduce fine-grained perturbations at the interface between the backbone and the feature enhancer, while preserving global feature statistics consistent with the clean representation. As a result, it improves robustness to target-domain variations without disrupting semantic information or the subsequent feature enhancement process.

\subsection{Training and Inference}

The prompt-driven image generation process is training-free, as the VLM generator is used solely to synthesize additional target-domain samples without parameter updates. After generation, each augmented sample is represented as \((\hat{I}, y, \hat{B})\) and combined with the original support set for detector fine-tuning. During fine-tuning, we update the image backbone, feature enhancer, and cross-modality decoder of GroundingDINO, while keeping the other components frozen. The optimization follows the original GroundingDINO objectives, including localization loss $Loss_l$ and classification loss $Loss_c$. Meanwhile, the NFP module is activated between the backbone and the feature enhancer to improve robustness to fine-grained target-domain variations.
During inference, both the VLM-based generation and noise perturbation are disabled. The model directly takes a test image and category text prompts as input and predicts bounding boxes using the standard GroundingDINO pipeline.
\section{Experiment}

\subsection{Experimental Setup}

\noindent{\textbf{Datasets.}}
Following the CD-ViTO benchmark~\cite{fu2024cross}, we evaluate our model on six cross-domain target datasets with diverse visual styles and domain characteristics: ArTaxOr~\cite{drange2019arthropod} for photorealistic arthropod images, Clipart1k~\cite{inoue2018cross} for cartoon-style illustrations, DIOR~\cite{li2020object} for aerial imagery, DeepFish~\cite{saleh2020realistic} and UODD~\cite{jiang2021underwater} for underwater scenarios, and NEU-DET~\cite{song2013noise} for industrial defect detection. We report results under the standard 1-shot, 5-shot, and 10-shot protocols.

\noindent{\textbf{Evaluation Metrics.}}
To verify the effectiveness of the proposed method, we adopt mean Average Precision (mAP, \%) as the primary evaluation metric. mAP is obtained by averaging AP over all categories. Following the COCO evaluation protocol~\cite{lin2014microsoft}, we compute the averaged result over IoU thresholds ranging from 0.50 to 0.95 with a step size of 0.05, thereby comprehensively evaluating the classification and localization performance of the model.

\noindent{\textbf{Implementation Details.}}
We use pretrained GroundingDINO~\cite{liu2024grounding} with Swin-Transformer Base (Swin-B)~\cite{liu2021swin} as the backbone and BERT-Base as the text encoder. The prompt-driven image generation stage is training-free. During fine-tuning, we update the image backbone, feature enhancer, and cross-modality decoder, while freezing the remaining modules. During training, we use the generated images and apply multi-scale resizing, random cropping, random horizontal flipping, and feature-level noise perturbation. We adopt AdamW with the learning rate and weight decay set to \(1\times10^{-4}\), and scale the backbone learning rate by 0.1. We evaluate the model under the COCO protocol without test-time augmentation. Dataset-specific settings, including noise strengths, training epochs, batch sizes, and learning rate decay milestones, are provided in the Appendix. All experiments are conducted on 8 NVIDIA L20X GPUs.

\begin{table}
    \centering
    \vspace{-0.05in}
    \caption{
        \textbf{Main results (mAP) on the CD-FSOD benchmark} under the 1/5/10-shot setting. 
       Best results are highlighted in pink.
    }    
    \vspace{-0.05in}
    \label{tab:main}
    \resizebox{1.\columnwidth}{!}{
    \begin{tabular}{cllcccccccc}
    \toprule
    & \textbf{Method} & \textbf{Backbone} & \textbf{ArTaxOr} & \textbf{Clipart1k} &  \textbf{DIOR} &   \textbf{DeepFish}   &  \textbf{NEU-DET}  & \textbf{UODD}  &  \textbf{Average} \\
    \midrule
    \multirow{12}{*}{\rotatebox{90}{1-shot}} & Meta-RCNN  ~\cite{yan2019meta} & ResNet50 & 2.8 & - & 7.8 & - & - & 3.6 & / \\
    & TFA w/cos ~\cite{wang2020frustratingly} & ResNet50 & 3.1 & - & 8.0 & - & - & 4.4 & / \\
    & FSCE ~\cite{sun2021fsce} & ResNet50 &  3.7 & - & 8.6 & - & - & 3.9 & / \\ 
    & DeFRCN ~\cite{qiao2021defrcn} & ResNet50 & 3.6 & - & 9.3 & - & - & 4.5  & / \\ 
    & Distill-CDFSOD ~\cite{xiong2023cd} & ResNet50 & 5.1 & 7.6 & 10.5 & nan &  nan &  5.9 & / \\
    \cline{2-10}
    & ViTDeT-FT ~\cite{li2022exploring} & ViT-B/14 & 5.9 & 6.1 & 12.9 & 0.9 & 2.4 & 4.0 & 5.4\\
    & Detic~\cite{zhou2022detecting} & ViT-L/14 & 0.6 & 11.4 & 0.1 & 0.9 & 0.0 & 0.0 & 2.2 \\
    & Detic-FT ~\cite{zhou2022detecting} & ViT-L/14 & 3.2 & 15.1 & 4.1 & 9.0 & 3.8 & 4.2 & 6.6 \\
    & DE-ViT~\cite{zhang2023detect} & ViT-L/14 & 0.4 & 0.5 & 2.7 & 0.4 & 0.4 & 1.5 & 1.0 \\
    & CD-ViTO~\cite{fu2024cross} & ViT-L/14 & 21.0 & 17.7 & 17.8 & 20.3 & 3.6 & 3.1 & 13.9 \\
    \cline{2-10} 
    & GroundingDINO$\dagger$~\cite{liu2024grounding} & Swin-B & 26.3 & 55.3 & 14.8 & 36.4 & 9.3 & 15.9 & 26.3 \\
    & ETS$\dagger$~\cite{pan2025enhance} & Swin-B & 28.1 & 55.8 & 12.7 & 39.3 & 11.7 & 18.9 & 27.8 \\
    & Domain-RAG~\cite{li2025domainrag} & Swin-B & 57.2 & 56.1 & 18.0 & 38.0 & 12.1 & 20.2 & 33.6 \\
    & \textbf{Ours (PSP-FSOD)} & Swin-B & \cellcolor{pink!60}\textbf{61.3} & \cellcolor{pink!60}\textbf{58.5} & \cellcolor{pink!60}\textbf{18.3} & \cellcolor{pink!60}\textbf{43.7} & \cellcolor{pink!60}\textbf{14.0} & \cellcolor{pink!60}\textbf{22.6} & \cellcolor{pink!60}\textbf{36.4} \\
    \midrule
    \multirow{12}{*}{\rotatebox{90}{5-shot}} 
    & Meta-RCNN  ~\cite{yan2019meta} & ResNet50 & 8.5 & - & 17.7 & - & - & 8.8 & / \\
    & TFA w/cos ~\cite{wang2020frustratingly} & ResNet50 & 8.8 & - & 18.1 & - & - & 8.7 & / \\
    & FSCE ~\cite{sun2021fsce} & ResNet50 & 10.2 & - & 18.7 & - & - & 9.6 & / \\ 
    & DeFRCN ~\cite{qiao2021defrcn} & ResNet50 &  9.9 & - & 18.9 & - & - & 9.9 & / \\ 
    & Distill-CDFSOD ~\cite{xiong2023cd} & ResNet50 & 12.5 & 23.3 & 19.1 & 15.5 & 16.0 & 12.2 & 16.4 \\
    \cline{2-10}
    & ViTDeT-FT ~\cite{li2022exploring} & ViT-B/14 & 20.9 & 23.3 & 23.3 & 9.0 & 13.5 & 11.1 & 16.9 \\
    & Detic~\cite{zhou2022detecting} & ViT-L/14 & 0.6 & 11.4 & 0.1 & 0.9 & 0.0 & 0.0 & 2.2 \\
    & Detic-FT ~\cite{zhou2022detecting} & ViT-L/14 & 8.7 & 20.2 & 12.1 & 14.3 & 14.1 & 10.4 & 13.3 \\
    & DE-ViT~\cite{zhang2023detect} & ViT-L/14 & 10.1 & 5.5 & 7.8 & 2.5 & 1.5 & 3.1 & 5.1 \\
    & CD-ViTO~\cite{fu2024cross} & ViT-L/14 & 47.9 & 41.1 & 26.9 & 22.3 & 11.4 & 6.8 & 26.1 \\
    \cline{2-10} 
    & GroundingDINO$\dagger$~\cite{liu2024grounding} & Swin-B & 68.4 & 57.6 & 29.6 & 41.6 & 19.7 & 25.6 & 40.4 \\
    & ETS$\dagger$~\cite{pan2025enhance} & Swin-B & 64.5 & 59.7 & 29.3 & 42.1 & 23.5 & 27.7 & 41.1 \\
    & Domain-RAG~\cite{li2025domainrag} & Swin-B & 70.0 & 59.8 & 31.5 & 43.8 & 24.2 & 26.8 & 42.7 \\
    & \textbf{Ours (PSP-FSOD)}  & Swin-B & \cellcolor{pink!60}\textbf{74.8} & \cellcolor{pink!60}\textbf{60.5} & \cellcolor{pink!60}\textbf{33.1} & \cellcolor{pink!60}\textbf{43.9} & \cellcolor{pink!60}\textbf{26.3} & \cellcolor{pink!60}\textbf{31.5} & \cellcolor{pink!60}\textbf{45.0} \\
    \midrule
    \multirow{12}{*}{\rotatebox{90}{10-shot}} 
    & Meta-RCNN  ~\cite{yan2019meta} & ResNet50 & 14.0 & - & 20.6 & - & - & 11.2 & / \\
    & TFA w/cos ~\cite{wang2020frustratingly} & ResNet50 & 14.8 & - & 20.5 & - & - & 11.8 & / \\
    & FSCE ~\cite{sun2021fsce} & ResNet50 & 15.9 & - & 21.9 & - & - & 12.0 & / \\ 
    & DeFRCN ~\cite{qiao2021defrcn} & ResNet50 & 15.5 & - & 22.9 & - & - & 12.1 & / \\ 
    & Distill-CDFSOD ~\cite{xiong2023cd} & ResNet50 & 18.1 & 27.3 & 26.5 & 15.5 & 21.1 & 14.5 & 20.5 \\ 
    \cline{2-10}
    & ViTDeT-FT ~\cite{li2022exploring} & ViT-B/14 & 23.4 & 25.6 & 29.4 & 6.5 & 15.8 & 15.6 & 19.4 \\ 
    & Detic~\cite{zhou2022detecting} & ViT-L/14 & 0.6 & 11.4 & 0.1 & 0.9 & 0.0 & 0.0 & 2.2 \\
    & Detic-FT ~\cite{zhou2022detecting} & ViT-L/14 & 12.0 & 22.3 & 15.4 & 17.9 & 16.8 & 14.4 & 16.5 \\
    & DE-ViT~\cite{zhang2023detect} & ViT-L/14 & 9.2 & 11.0 & 8.4 & 2.1 & 1.8 & 3.1 & 5.9 \\
    & CD-ViTO~\cite{fu2024cross} & ViT-L/14 & 60.5 & 44.3 & 30.8 & 22.3 & 12.8 & 7.0 & 29.6 \\
    \cline{2-10} 
    & GroundingDINO$\dagger$~\cite{liu2024grounding} & Swin-B & 73.0 & 58.6 & 37.2 & 38.5 & 25.5 & 30.3 & 43.9 \\
    & ETS$\dagger$~\cite{pan2025enhance} & Swin-B & 70.6 & 60.8 & 37.5 & 42.8 & 26.1 & 28.3 & 44.4 \\
    & Domain-RAG~\cite{li2025domainrag} & Swin-B & 73.4 & 61.1 & 39.0 & 41.3 & 26.3 & 31.2 & 45.4 \\
    & \textbf{Ours (PSP-FSOD)} & Swin-B & \cellcolor{pink!60}\textbf{75.3} & \cellcolor{pink!60}\textbf{62.0} & \cellcolor{pink!60}\textbf{39.5} & \cellcolor{pink!60}\textbf{45.7} & \cellcolor{pink!60}\textbf{27.8} & \cellcolor{pink!60}\textbf{32.4} & \cellcolor{pink!60}\textbf{47.1} \\
    \bottomrule
    \end{tabular}}
    \vspace{-0.2in}
\end{table}

\subsection{Comparison to Competitive Approaches}

We compare our method with a diverse set of state-of-the-art approaches, including Meta-RCNN, TFA w/cos, FSCE, DeFRCN, Distill-CDFSOD, ViTDet-FT, Detic/Detic-FT, DE-ViT, and CD-ViTO, partially following CD-ViTO for reported results. In particular, ETS and Domain-RAG are also based on data augmentation.
Table~\ref{tab:main} summarizes the main results of our method on the CD-FSOD dataset across six novel target domains under the 1/5/10-shot settings. As shown in the table, our method consistently outperforms existing competitors on most target domains, achieving the best average mAP under all settings. Compared with the fine-tuned GroundingDINO baseline, our method improves average mAP by 10.1, 4.6, and 3.2 points under the 1-, 5-, and 10-shot settings, respectively, demonstrating the effectiveness of our generated target-domain samples and feature-level noise perturbation in improving detector robustness and adaptation.
Compared with other data augmentation-based methods, our method also outperforms ETS by 8.6, 3.9, and 2.7 points, and Domain-RAG by 2.8, 2.3, and 1.7 points under the 1-, 5-, and 10-shot settings, respectively, further validating the effectiveness of our PDS module over existing augmentation strategies.

Beyond the average gains, we observe three consistent trends.
\textbf{1) Strong performance under low-shot settings.} Our method achieves 36.4 and 45.0 mAP under the 1- and 5-shot settings, outperforming GroundingDINO by 10.1 and 4.6 points, ETS by 8.6 and 3.9 points, and Domain-RAG by 2.8 and 2.3 points, respectively, indicating improved sample efficiency under scarce supervision.
\textbf{2) Larger gains on complex domains.} More pronounced improvements are observed on ArTaxOr, DeepFish, UODD, and NEU-DET, where stronger foreground variation and foreground-background coupling make adaptation more challenging.
\textbf{3) Limited gains on simpler domains.} Improvements on Clipart and DIOR are relatively smaller, where domain shifts are less severe.

\begin{figure}[h]
    \centering
    \includegraphics[width=\linewidth]{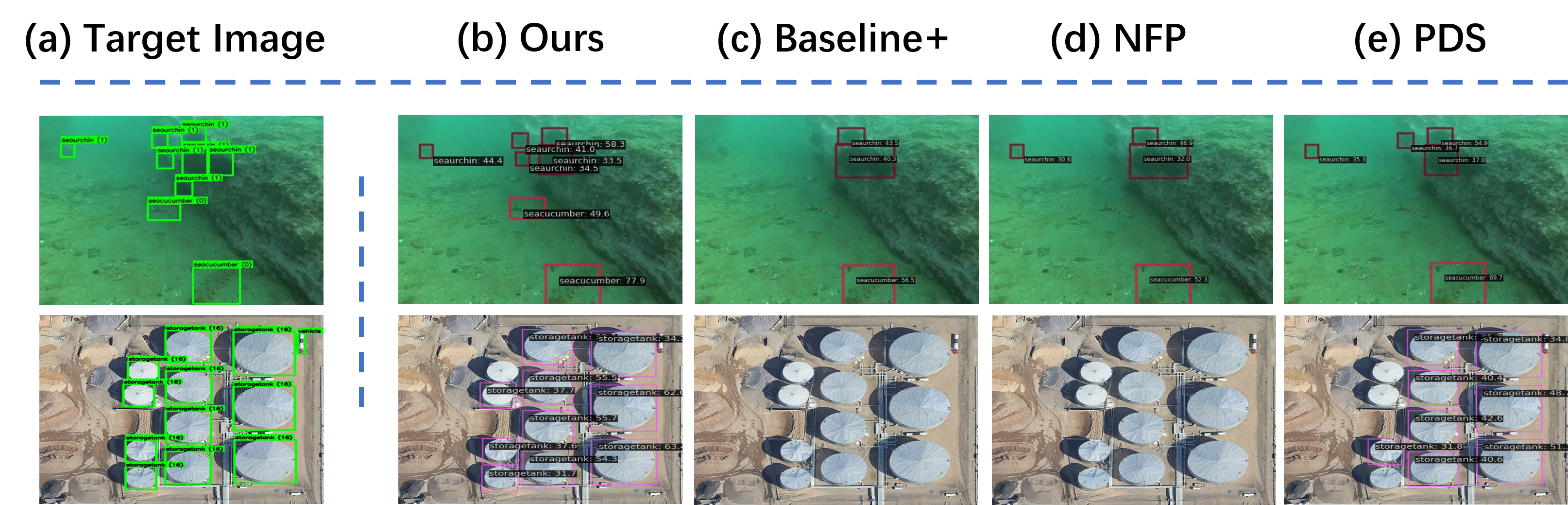}
    \vspace{-0.22in}
    \caption{Visualization of different components on target-domain images under the 1-shot setting.}
    \vspace{-0.02in}
    \label{fig:ablation_dect}
\end{figure}

\subsection{Model Analysis}

\begin{table}[h]
    \centering
    \vspace{-0.05in}
    \caption{\textbf{Ablation results (mAP) on CD-FSOD} under the 1/5/10-shot settings. ``Noise'' denotes feature-level perturbation with mean-variance restoration, while ``$\textbf{I}_{\text{gen}}$'' denotes PDS-generated target-domain samples. A checkmark indicates that the corresponding module is used.}
    \vspace{-0.05in}
    \label{tab:ablation_noise_genfig}
    \begin{adjustbox}{max width=\textwidth}
    \begin{tabular}{clccccccccc}
    \toprule
    \textbf{Shot} 
    & \textbf{Method}
    & \textbf{Noise}
    & $\textbf{I}_{gen}$
    & \textbf{ArTaxOr} 
    & \textbf{Clipart1k} 
    & \textbf{DIOR} 
    & \textbf{DeepFish} 
    & \textbf{NEU-DET} 
    & \textbf{UODD} 
    & \textbf{Average} \\
    \midrule

    \multirow{4}{*}{\rotatebox{90}{1-shot}}
    & Baseline+ &  &  & 26.8 & 55.8 & 14.6 & 42.2 & 9.9  & 17.5 & 27.8 \\
    & NFP & \checkmark &  & 34.0 & 57.0 & 15.1 & 36.7 & 10.2 & 20.1 & 28.9 \\
    & PDS &  & \checkmark & 58.3 & 57.6 & 16.2 & 42.6 & 12.6 & 17.9 & 34.2 \\
    & \textbf{Ours} & \checkmark & \checkmark 
        & \cellcolor{pink!60}\textbf{61.3} 
        & \cellcolor{pink!60}\textbf{58.5} 
        & \cellcolor{pink!60}\textbf{18.3} 
        & \cellcolor{pink!60}\textbf{43.7} 
        & \cellcolor{pink!60}\textbf{14.0} 
        & \cellcolor{pink!60}\textbf{22.6} 
        & \cellcolor{pink!60}\textbf{36.4} \\
    \midrule

    \multirow{4}{*}{\rotatebox{90}{5-shot}}
    & Baseline+ &  &  & 70.0 & 57.9 & 27.9 & 42.3 & 20.6 & 27.7 & 41.1 \\
    & NFP & \checkmark &  & 73.1 & 58.7 & 26.9 & 42.0 & 22.7 & 26.7 & 41.7 \\
    & PDS &  & \checkmark & 74.2 & 60.2 & 32.9 & 42.5 & 25.7 & 30.3 & 44.3 \\
    & \textbf{Ours} & \checkmark & \checkmark 
        & \cellcolor{pink!60}\textbf{74.8} 
        & \cellcolor{pink!60}\textbf{60.5} 
        & \cellcolor{pink!60}\textbf{33.1} 
        & \cellcolor{pink!60}\textbf{43.9} 
        & \cellcolor{pink!60}\textbf{26.3} 
        & \cellcolor{pink!60}\textbf{31.5} 
        & \cellcolor{pink!60}\textbf{45.0} \\
    \midrule

    \multirow{4}{*}{\rotatebox{90}{10-shot}}
    & Baseline+ &  &  & 73.3 & 59.1 & 36.4 & 40.7 & 25.5 & 31.7 & 44.5 \\
    & NFP & \checkmark &  & 73.6 & 61.2 & 37.3 & 42.2 & 25.7 & 32.1 & 45.4 \\
    & PDS &  & \checkmark 
        & 74.7 
        & 61.8 
        & 39.1 
        & 43.8 
        & \cellcolor{pink!60}\textbf{28.3} 
        & 31.9 
        & 46.6 \\
    & \textbf{Ours} & \checkmark & \checkmark 
        & \cellcolor{pink!60}\textbf{75.3} 
        & \cellcolor{pink!60}\textbf{62.0} 
        & \cellcolor{pink!60}\textbf{39.5} 
        & \cellcolor{pink!60}\textbf{45.7} 
        & 27.8 
        & \cellcolor{pink!60}\textbf{32.4} 
        & \cellcolor{pink!60}\textbf{47.1} \\
    \bottomrule
    \end{tabular}
    \end{adjustbox}
    \vspace{-0.2in}
\end{table}



This section analyzes different components. For fair comparison, GroundingDINO fine-tuned with our training strategy is used as \textbf{Baseline+}. We then add Prompt-Driven Domain Simulation (PDS) and Noise-Induced Feature Perturbation (NFP) to evaluate their individual and combined effects.

Table~\ref{tab:ablation_noise_genfig} shows that both modules consistently improve performance, and their combination yields further gains. Compared with \textbf{Baseline+}, feature-level noise perturbation improves the average mAP by +1.1, +0.6, and +0.9 under the 1/5/10-shot settings, while generated figures bring larger gains of +6.4, +3.2, and +2.1. 
When combined, the performance is further boosted to +8.6, +3.9, and +2.6, indicating that generated figures are the dominant factor, particularly in the low-shot regime.
This highlights their strong complementarity: generated figures primarily expand structural diversity, whereas feature-level noise perturbation acts as a regularizer that mitigates residual synthetic-to-real discrepancies and improves feature robustness. 


Fig.~\ref{fig:ablation_dect} provides visualization results that further validate the effectiveness of PSP-FSOD. It demonstrates that PSP-FSOD consistently achieves accurate and complete detections in target-domain images with cluttered backgrounds and large appearance variations. By integrating prompt-driven domain simulation with feature-level perturbation, our method enhances robustness to complex scenes and object shape changes.
From a dataset perspective, generated figures are more effective for datasets with strong foreground variations and complex foreground–background interactions (e.g., UODD), while feature-level noise perturbation better handles cases with blurred boundaries, texture disturbances, or unstable imaging quality (e.g., DeepFish and NEU-DET). Their complementary strengths enable more stable improvements across diverse challenging target domains.



\subsection{Ablation Study}
\label{sec:prompt_simulation}

\paragraph{Influence of Various Data Augmentation Strategies.} 
Table~\ref{tab:generation_strategy_ablation} compares our proposed PDS-based foreground–background adaptation with different data augmentation strategies under the 1-shot setting. Our method consistently outperforms all baselines across six benchmarks for cross-domain few-shot object detection. 
In particular, compared with the previously best-performing background-adaptation strategy, our foreground–background adaptation, enabled by prompt-driven domain simulation, achieves substantial gains of +4.6 (ArTaxOr), +1.5 (Clipart1k), +0.9 (DIOR), +8.4 (DeepFish), +5.1 (NEU-DET), and +1.0 (UODD).
This validates the importance of jointly modeling object-level and scene-level variations, confirming that integrating foreground and background adaptation yields the most effective target-domain augmentation under extremely limited supervision.

\begin{table}[h]
    \centering
    \vspace{-0.05in}
    \caption{
        \textbf{Ablation results on different data augmentation strategies} under the 1-shot setting.
    }
    \vspace{-0.05in}
    \label{tab:generation_strategy_ablation}
    \begin{adjustbox}{max width=\textwidth}
    \begin{tabular}{lcccccc}
    \toprule
    \textbf{Method} 
    & \textbf{ArTaxOr} 
    & \textbf{Clipart1k} 
    & \textbf{DIOR} 
    & \textbf{DeepFish} 
    & \textbf{NEU-DET} 
    & \textbf{UODD} \\
    \midrule
    GroundingDINO~\cite{liu2024grounding}      & 26.3 & 55.3 & 14.8 & 36.4 & 9.3  & 15.9 \\
    Copy-Paste         & 38.8 & 55.0 & 15.0 & 36.4 & 8.4  & 14.2 \\
    Foreground Adaptation  & 29.9 & 56.7   & 15.1   & 37.9 & 8.6  & 16.2 \\
    Background Adaptation  & 53.7 & 56.1   & 15.3   & 34.2 & 7.5  & 16.9 \\
    \cellcolor{pink!60}
    Foreground-Background Adaptation   
        & \cellcolor{pink!60}\textbf{58.3} 
        & \cellcolor{pink!60}\textbf{57.6} 
        & \cellcolor{pink!60}\textbf{16.2} 
        & \cellcolor{pink!60}\textbf{42.6} 
        & \cellcolor{pink!60}\textbf{12.6} 
        & \cellcolor{pink!60}\textbf{17.9} \\
    \bottomrule
    \end{tabular}
    \end{adjustbox}
    \vspace{-0.2in}
\end{table}

\paragraph{\textbf{Comparison of Different Prompt-Driven Domain Simulation.}} 
Table~\ref{tab:artaxor_prompt_ablation} compares the performance of our proposed PSP-FSOD on ArTaxOr when trained with different types of samples generated by 

\vspace{-0.07in}
\begin{wraptable}{r}{0.52\textwidth}
    \centering
    \vspace{-0.05in}
    \caption{\textbf{Ablation of various prompt-driven generation strategies on ArTaxOr.} 
        ``$\textbf{PDS}_{fg}$'' and ``$\textbf{PDS}_{bg}$'' denote foreground-only and background-only adaptations, respectively, under a consistent foreground–background generation schema.}
    \vspace{-0.05in}
    \label{tab:artaxor_prompt_ablation}
    \begin{tabular}{lccc}
    \toprule
    \textbf{Strategy}      &\textbf{1-shot} &\textbf{5-shot} &\textbf{10-shot} \\
    \midrule
    Inconsistent Bg-Fg      &48.5 &74.2 &73.6 \\
    $\textbf{PDS}_{bg}$     &54.2 &71.4 &71.1 \\
    $\textbf{PDS}_{fg}$    &43.1 &70.6 &\cellcolor{pink!60}\textbf{75.9} \\
    \textbf{Ours (PSP-FSOD)}
        &\cellcolor{pink!60}\textbf{61.3}
        &\cellcolor{pink!60}\textbf{74.8}
        &75.3 \\
    \bottomrule
    \end{tabular}
    \vspace{-0.05in}
\end{wraptable}
 various prompt-driven generation strategies.
The results show that the proposed PSP-FSOD achieves the best performance under the 1-shot and 5-shot settings, reaching 61.3 and 74.8 mAP, respectively, clearly outperforming other prompt designs.
Specifically, compared with synthetic data exhibiting inconsistent foreground–background semantics(48.5/74.2/73.6 in 1/5/10-shot), the three settings with consistent pairing achieve substantially better performance. Under this consistent setting, background-only modification outperforms foreground-only modification in most cases (54.2 vs. 43.1 in 1-shot, 71.4 vs. 70.6 in 5-shot), while jointly adapting both yields the best results. This further confirms the effectiveness of simultaneous foreground–background adaptation for CD-FSOD.

\paragraph{Hyperparameters of Noise-Induced Feature Perturbation.} 
\begin{wraptable}{r}{0.65\textwidth}
    \centering
    \caption{\textbf{Effect of different noise strengths on different feature levels for NFP under UODD.} L1--L4 denote different feature levels; R denotes distribution rectification.}
    \vspace{-0.05in}
    \label{tab:noise_ablation}
    \begin{tabular}{cccccccc}
    \toprule
    \textbf{L1} & \textbf{L2} & \textbf{L3} & \textbf{L4} & \textbf{R} & \textbf{1-shot} & \textbf{5-shot} & \textbf{10-shot} \\
    \midrule
    0    & 0    & 0    & 0    & \cmark & 17.9 & 30.3 & 31.7 \\
    0.15 & 0    & 0    & 0    & \cmark & 21.8 & 29.0 & 32.3 \\
    0.15 & 0.15 & 0    & 0    & \cmark & 20.2 & 28.3 & 31.9 \\
    \rowcolor{gray!25}
    0.15 & 0.15 & 0.15 & 0    & \cmark & 19.9 & 28.1 & 31.4 \\
    \rowcolor{gray!25}
    0.15 & 0.15 & 0.1  & 0    & \cmark & 22.3 & 29.1 & 32.0 \\
    0.15 & 0.15 & 0.1  & 0.1  & \xmark & 19.5 & 29.6 & 31.9 \\
    \rowcolor{pink!60}
    \textbf{0.15} & \textbf{0.15} & \textbf{0.1} & \textbf{0.1} & \textbf{\cmark} & \textbf{22.6} & \textbf{31.5} & \textbf{32.4} \\
    \bottomrule
    \end{tabular}
    \vspace{-0.05in}
\end{wraptable}
Table~\ref{tab:noise_ablation} ablates the effect of different noise magnitudes applied to different feature levels under the UODD setting. Overall, compared with the first six rows, introducing NFP at each feature level consistently improves performance, demonstrating the effectiveness of noise perturbation across hierarchical representations. Comparing the two gray rows, we find that stronger perturbation on low-level features and weaker perturbation on high-level features yields better performance (19.9/28.1/31.4 vs. 22.3/29.1/32.0). Furthermore, comparing the last two rows shows that applying distribution rectification consistently improves performance, indicating that preserving feature statistics helps stabilize the effect of noise perturbation. The best setting, “0.15 0.15 0.1 0.1”, achieves the highest mAP with a +4.7 gain over the baseline.

\subsection{Conclusion}
This work presents PSP-FSOD, a prompt-driven domain simulation framework with feature perturbation regularization for cross-domain few-shot object detection. 
It leverages VLM-based generation to synthesize semantically consistent and domain-diverse samples via joint foreground–background modeling, and further enhances robustness through noise-induced feature perturbation with distribution correction, which stabilizes training and reduces reliance on domain-specific cues.
Extensive experiments on multiple CD-FSOD benchmarks demonstrate consistent gains over existing methods. Overall, PSP-FSOD highlights the effectiveness of combining controllable prompt-driven data synthesis with feature-level regularization for cross-domain few-shot learning.
Moreover, the proposed framework is general and readily extends to other few-shot learning tasks.

{\small
\bibliographystyle{unsrtnat}
\bibliography{neurips_2026}
}

\appendix

\appendix
\section{More Implementation Details and Analysis}
This section presents additional implementation details of the proposed CD-FSOD method and comparison of the number of the generated samples.

\subsection{More Implementation Details}
Additional implementation details are summarized in Table~\ref{tab:training_hyperparams}, including dataset-specific epochs, batch sizes, and learning rate decay milestones. 
Other hyperparameters are kept unchanged across datasets. 
Random horizontal flipping is applied with a probability of 0.5, and feature-level noise perturbation is applied in every forward pass. 
For image synthesis, two augmented images are generated per support image on ArTaxOr~\cite{drange2019arthropod}, while one is generated on the other datasets. 
The learning rate is decayed by 0.1 using MultiStepLR at the specified milestones.
\begin{table}[h]
    \centering
    \vspace{-0.05in}
    \caption{
        \textbf{Training hyper-parameters for different datasets.}
    }
    \vspace{-0.05in}
    \label{tab:training_hyperparams}
    \setlength{\tabcolsep}{18pt}
    \renewcommand{\arraystretch}{1.12}
    \resizebox{0.85\columnwidth}{!}{
    \begin{tabular}{lccc}
    \toprule
    \textbf{Dataset} & \textbf{Epochs} & \textbf{Batch size} & \textbf{LR decay epochs} \\
    \midrule
    ArTaxOr~\cite{drange2019arthropod}   & 60 & 4 & 10 \\
    Clipart1k~\cite{inoue2018cross}      & 30 & 3 & 2, 8 \\
    DIOR~\cite{li2020object}             & 80 & 4 & 5 \\
    DeepFish~\cite{saleh2020realistic}   & 30 & 4 & 5 \\
    NEU-DET~\cite{song2013noise}         & 60 & 4 & 5 \\
    UODD~\cite{jiang2021underwater}      & 60 & 4 & 5 \\
    \bottomrule
    \end{tabular}}
    \vspace{-0.1in}
\end{table}

\subsection{More Analysis}
\paragraph{Comparison of the Number of Generated Image.}
\begin{table}[h]
\centering
\caption{
\textbf{Comparison of the number of generated images per support image.}
}
\label{tab:gen_num_compare}
\setlength{\tabcolsep}{8pt}
\resizebox{0.9\linewidth}{!}{
\begin{tabular}{lcc}
\toprule
\textbf{Method} & \textbf{Dataset} & \textbf{Generated images per support image} \\
\midrule
Domain-RAG~\cite{li2025domainrag} & All datasets & 5 \\
\midrule
\multirow{2}{*}{Ours} & ArTaxOr~\cite{drange2019arthropod} & 2 \\
                      & Other datasets & 1 \\
\bottomrule
\end{tabular}
}
\end{table}
The number of generated images per support image is compared with Domain-RAG in Table~\ref{tab:gen_num_compare}. Domain-RAG generates five synthetic images for each support image, while the proposed method generates only one image on most datasets and two images on ArTaxOr. The slightly larger number on ArTaxOr is used to cover more diverse insect shapes, poses, and textures. This shows that, under the semantic consistency constraints, even a minimal number of generated images can provide effective target-domain variations.

\section{More Visualization for PSP-FSOD}
In this section, we provide additional visual results, including the prompts used for domain simulation, more qualitative detection results, analyses of the generated samples, and representative failure cases.

\subsection{Visualization of Different Prompt-Driven Domain Simulation}
\begin{figure}[h]
    \centering
    \includegraphics[width=\linewidth]{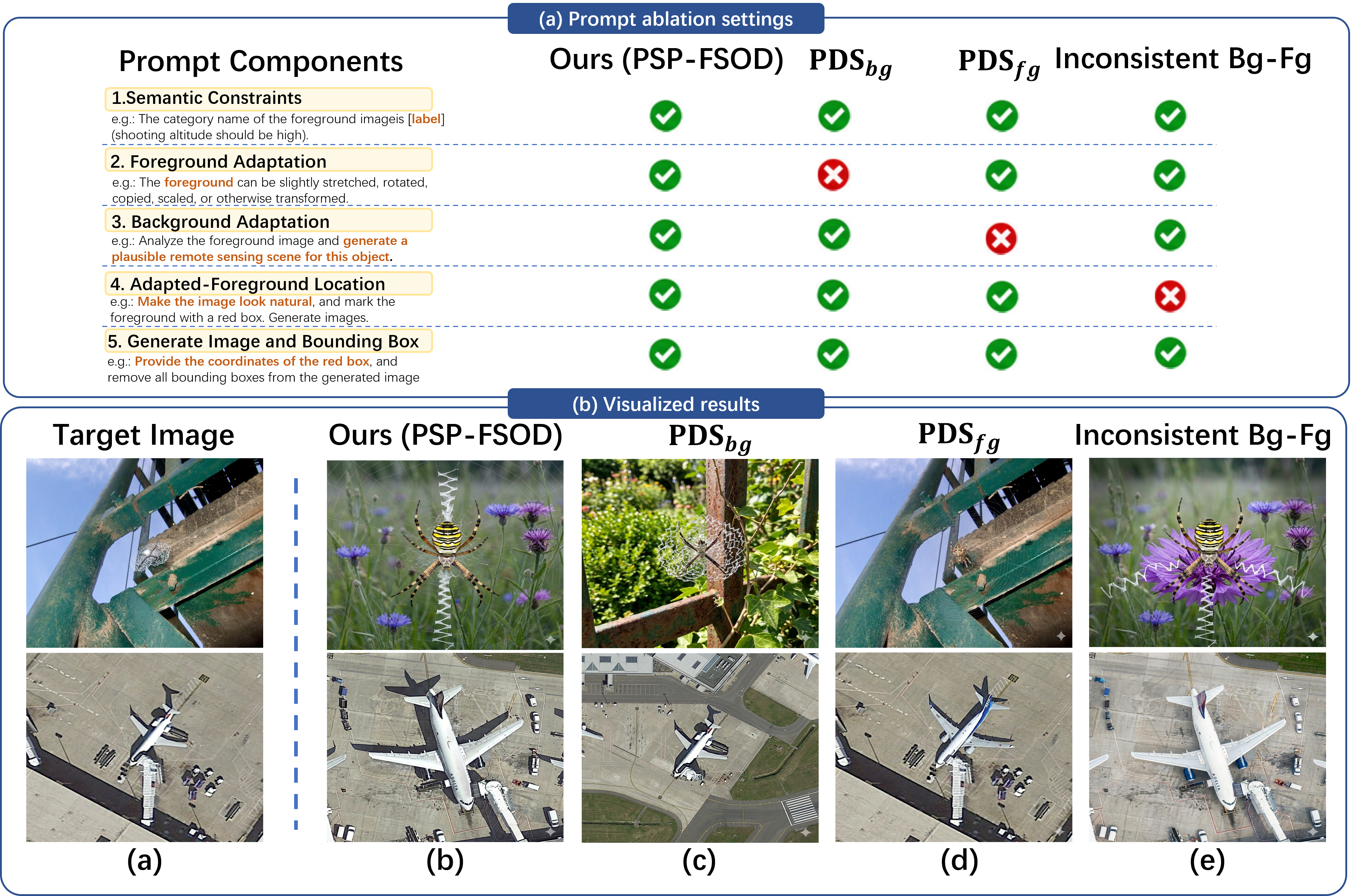}
    \caption{
    Visualization comparison of different prompt-driven domain simulation strategies. }
    \label{fig:a_vis_prompt}
\end{figure}
Fig.~\ref{fig:a_vis_prompt} provides a visual comparison of domain samples generated by the VLM using different prompt designs, illustrating the effect of each prompt component on domain simulation.
``Inconsistent Bg-Fg'' represents prompts that do not enforce semantic consistency between foreground and background generation. As a result, it may produce unnatural compositions, such as the airplane without reasonable shadows and the spider mismatched with the flower background shown in Fig.~\ref{fig:a_vis_prompt}(e).
``$\textbf{PDS}_{fg}$'' mainly modifies the foreground while keeping the background close to the input image. During prompt-controlled VLM generation, semantic consistency between foreground and background is preserved, while only the foreground appearance, such as the spider category or airplane style, is changed, as shown in Fig.~\ref{fig:a_vis_prompt}(d).
``$\textbf{PDS}_{bg}$'' controls only the background variations while keeping the foreground consistent. As shown in Fig.~\ref{fig:a_vis_prompt}(c), the vegetation scene for the spider and the remote-sensing scene for the airplane are largely preserved, while the foreground changes remain limited.
These two variants are insufficient for comprehensive domain simulation, resulting in suboptimal models trained with them.
In contrast, the PDS-based augmentation strategy of our proposed PSP-FSOD jointly accounts for both foreground and background variations while maintaining semantic consistency, enabling more comprehensive domain simulation and better adaptation to target-domain characteristics.
\subsection{Qualitative Comparison of Detection Results}
\begin{figure}[h]
    \centering
    \includegraphics[width=\linewidth]{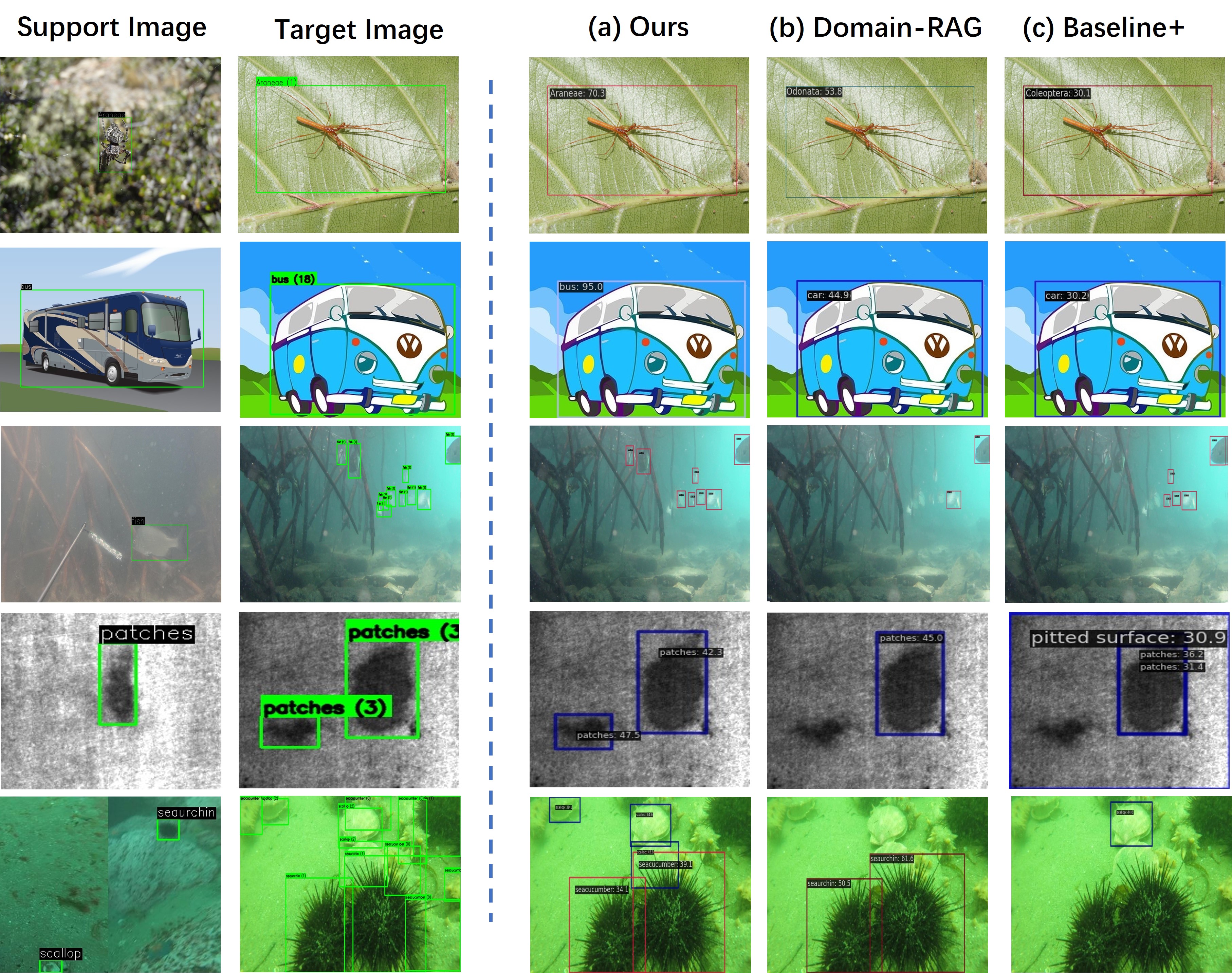}
    \caption{
    Visualization comparison of different methods on target-domain images. “Baseline+” is consistent with the Baseline+ setting reported in Table~\ref{tab:ablation_noise_genfig} . Our method achieves more accurate localization and classification under foreground shape variations and complex background interference.}
    \label{fig:visul_compare}
\end{figure}
Fig.~\ref{fig:visul_compare} presents qualitative comparisons on different target-domain images. Our method achieves more reliable detections when the target objects differ noticeably from the support images in appearance, pose, or shape. As shown in the first two rows, it can not only localize the objects accurately but also predict the correct categories, whereas Domain-RAG and Baseline+ are more prone to misclassification or low-confidence predictions. This suggests that our method improves recognition by better covering foreground shape variations.

Our method also shows stronger generalization in challenging scenes with complex backgrounds and blurred foreground-background boundaries. As shown in rows 3--5, it can still localize targets effectively in underwater and industrial images, where background interference is strong and object boundaries are unclear. This benefit mainly comes from the combination of background variation simulation and noise-induced feature perturbation, which improves robustness to boundary ambiguity and local visual disturbances.

\subsection{More Detection Results of PSP-FSOD}
\begin{figure}[h]
    \centering
    \includegraphics[width=\linewidth]{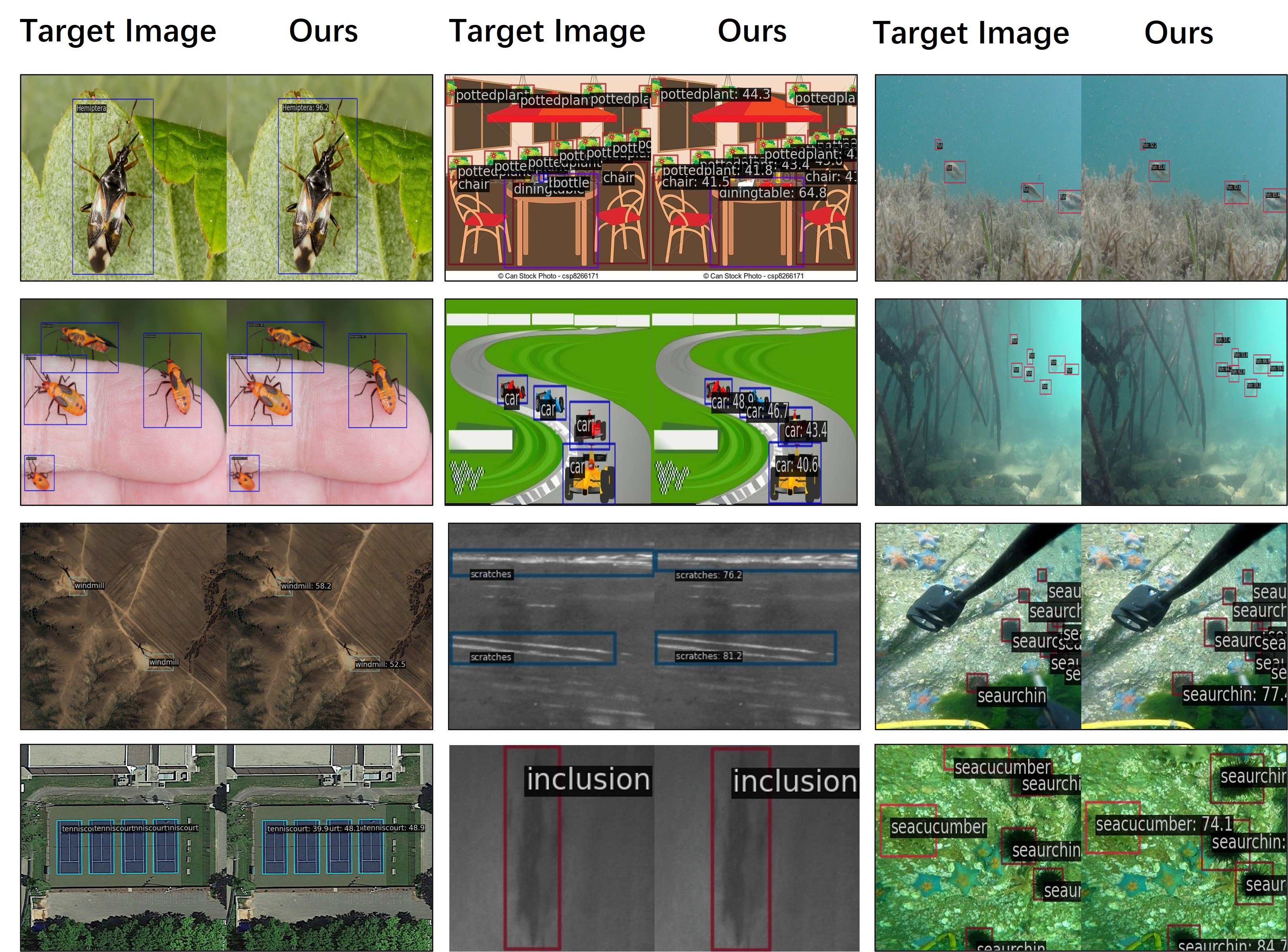}
    \caption{
    Additional detection results of PSP-FSOD on diverse target-domain images.}
    \label{fig:a_more_dect}
\end{figure}
Fig.~\ref{fig:a_more_dect} presents additional detection results of PSP-FSOD on diverse target-domain images. By generating diverse foreground-background combinations, PSP-FSOD better recognizes objects with different appearances and shapes, showing strong adaptation to domains with large foreground variations, such as ArTaxOr. Combined with noise-induced feature perturbation, it also improves robustness to blurred, low-contrast, and visually degraded industrial or underwater scenes.

\subsection{More Generation Results of PSP-FSOD}
\begin{figure}[h]
    \centering
    \includegraphics[width=\linewidth]{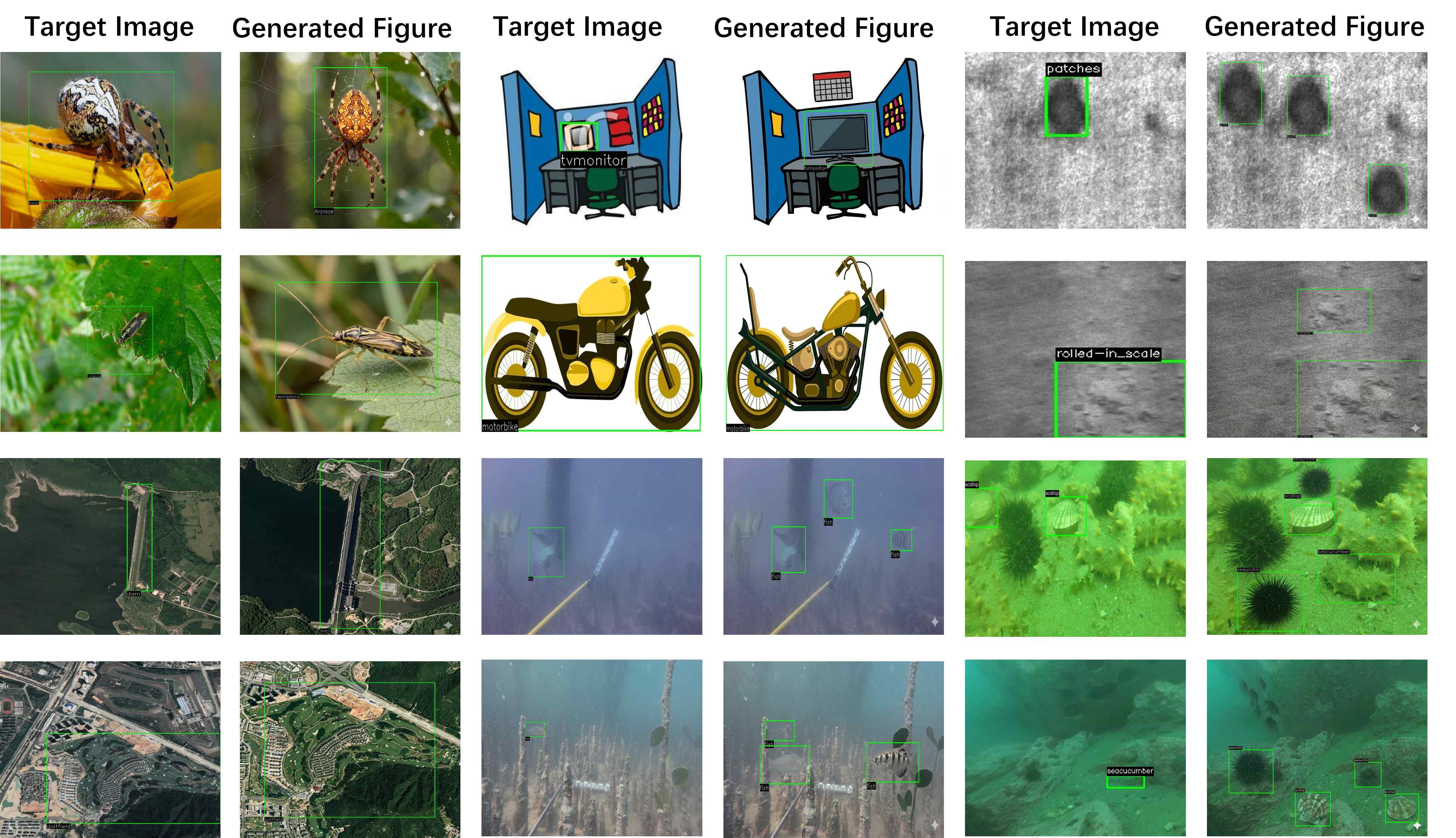}
    \caption{
Additional generation results of PSP-FSOD. Bounding boxes are overlaid only for visualization, showing that the generated images are semantically coherent and well localized.}
    \label{fig:more_generation}
\end{figure}

\begin{figure}[ht]
    \centering
    \includegraphics[width=\linewidth]{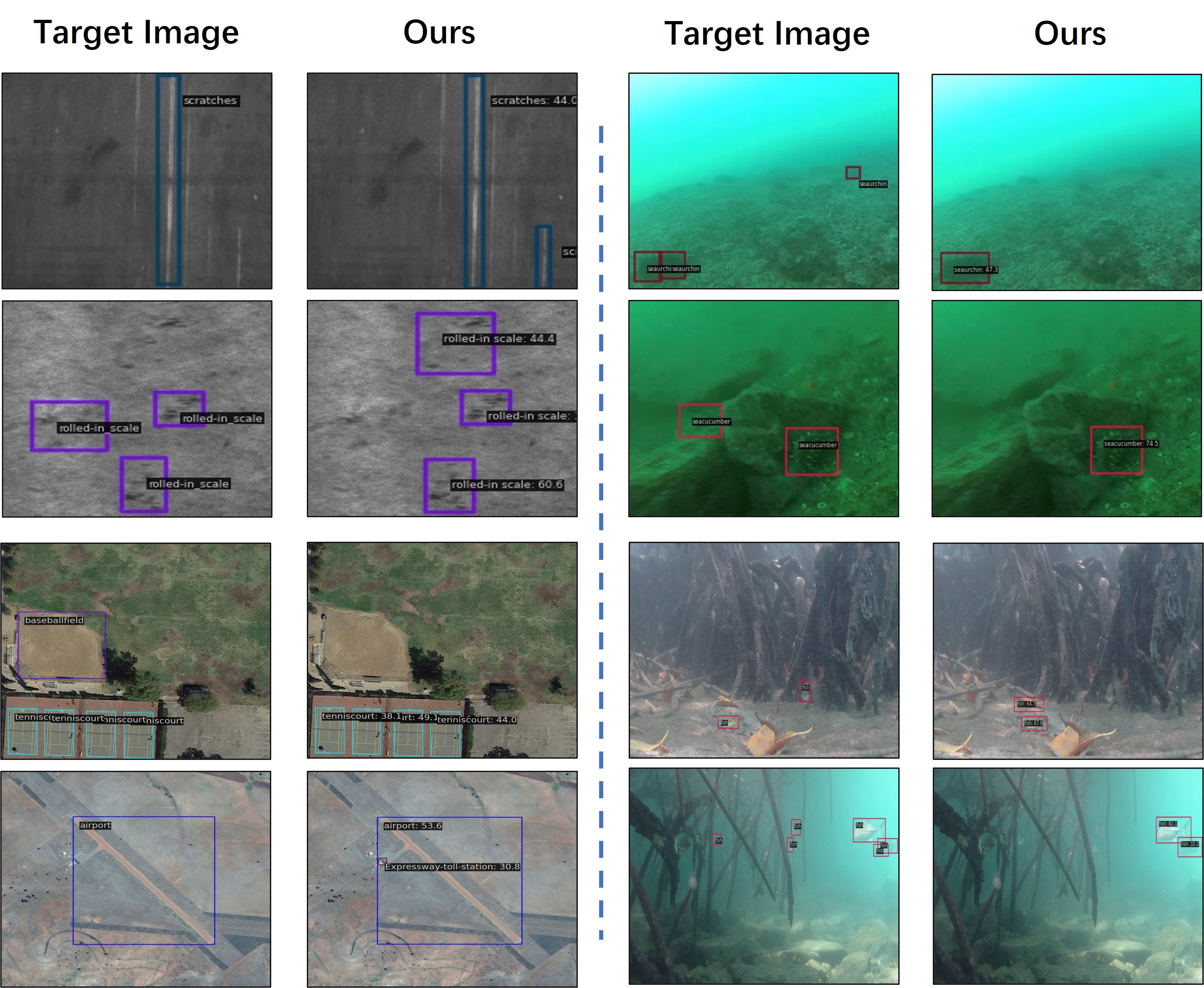}
    \caption{
    Failure cases of PSP-FSOD on challenging target-domain images with severe domain shifts, weak visibility, small objects, and ambiguous backgrounds.}
    \label{fig:a_failure}
\end{figure}
Fig.~\ref{fig:more_generation} presents additional generation results of PSP-FSOD on diverse target-domain images. The generated samples exhibit clear variations in both object shape and background context, while maintaining harmonious foreground-background composition. In addition, the generated object regions remain well aligned, without obvious distortion or skewed bounding boxes. These results show that PSP-FSOD can synthesize visually coherent and semantically plausible images with diverse foreground and background changes.

\subsection{Failure Cases}
\label{failure}
Fig.~\ref{fig:a_failure} shows typical failure cases of PSP-FSOD. The errors mainly occur in challenging scenarios with severe domain shifts, small or low-visibility objects, low contrast, and cluttered backgrounds, indicating that detecting extremely subtle or visually degraded objects remains difficult and remains challenging for the underlying GroundingDINO detector.

\section{Analysis on Limitations and Future Work}
In addition to the failure cases under severe domain shifts, weak visibility, dense small objects, and ambiguous backgrounds discussed in Sec.~\ref{failure}, PSP-FSOD still has two limitations. First, the prompt templates are manually designed and may require domain-specific prior knowledge. Second, the noise strengths for different feature levels are manually selected, which may not be optimal across all domains. Future work will explore automatic agent-based prompt optimization and adaptive noise strength learning to reduce manual design, while further improving robustness under severe domain shifts through small-object-aware enhancement, hard-example mining, and stronger domain-invariant representation learning.

\section{Broader Impacts}
This work shows that semantically consistent VLM-generated images can provide useful supervision for few-shot object detection, improving generalization under limited annotation and cross-domain shifts. Moreover, VLM-based image generation and feature-level noise perturbation show complementary effects: generated images expand structural diversity, while feature-level noise perturbation regularizes residual synthetic-to-real gaps and improves feature robustness. Although the prompt templates still require manual design, our results provide an effective first step toward VLM-driven data synthesis for CD-FSOD, which can be further extended with agent-based automatic prompt design in future work.

\clearpage
\section*{NeurIPS Paper Checklist}

\begin{enumerate}

\item {\bf Claims}
    \item[] Question: Do the main claims made in the abstract and introduction accurately reflect the paper's contributions and scope?
    \item[] Answer: \answerYes{} 
    \item[] Justification: The main claims in the abstract and introduction accurately reflect the proposed method, experimental setting, and reported results. The contributions and empirical evidence are described in the introduction and experimental sections.

\item {\bf Limitations}
    \item[] Question: Does the paper discuss the limitations of the work performed by the authors?
    \item[] Answer: \answerYes{} 
    \item[] Justification: The paper discusses limitations of the proposed method in the limitations section of Appendix.

\item {\bf Theory assumptions and proofs}
    \item[] Question: For each theoretical result, does the paper provide the full set of assumptions and a complete (and correct) proof?
    \item[] Answer: \answerNA{} 
    \item[] Justification: This paper does not present theoretical results or formal proofs. The proposed method is evaluated empirically through experiments and ablation studies.

    \item {\bf Experimental result reproducibility}
    \item[] Question: Does the paper fully disclose all the information needed to reproduce the main experimental results of the paper to the extent that it affects the main claims and/or conclusions of the paper (regardless of whether the code and data are provided or not)?
    \item[] Answer: \answerYes{} 
    \item[] Justification: The paper provides the experimental settings needed to reproduce the main results, including datasets, evaluation protocols, implementation details, training configurations, and ablation settings. The model components and hyperparameters are described in the method and experimental sections, with additional details provided in the appendix.

\item {\bf Open access to data and code}
    \item[] Question: Does the paper provide open access to the data and code, with sufficient instructions to faithfully reproduce the main experimental results, as described in supplemental material?
    \item[] Answer: \answerYes{} 
    \item[] Justification: An anonymized code repository is provided with instructions for reproducing the main experiments. The paper also describes the datasets, evaluation protocols, and implementation details, and all datasets used in the experiments are publicly available.

\item {\bf Experimental setting/details}
    \item[] Question: Does the paper specify all the training and test details (e.g., data splits, hyperparameters, how they were chosen, type of optimizer) necessary to understand the results?
    \item[] Answer: \answerYes{} 
    \item[] Justification: The paper specifies the experimental settings, including datasets, few-shot splits, training and testing protocols, evaluation metrics, optimizer, learning rate, batch size, and other key hyperparameters. Additional implementation details are provided in the appendix. We will release the source code to provide a clear path for reproducing the reported results.

\item {\bf Experiment statistical significance}
    \item[] Question: Does the paper report error bars suitably and correctly defined or other appropriate information about the statistical significance of the experiments?
    \item[] Answer: \answerNo{} 
    \item[] Justification: The paper reports mAP results under the standard CD-FSOD evaluation protocol, but does not include error bars, confidence intervals, or statistical significance tests. Running multiple seeds for all datasets and shot settings is computationally expensive due to the VLM-based image synthesis and detector training pipeline. We acknowledge this as a limitation and will include multi-run statistical analysis in future work.

\item {\bf Experiments compute resources}
    \item[] Question: For each experiment, does the paper provide sufficient information on the computer resources (type of compute workers, memory, time of execution) needed to reproduce the experiments?
    \item[] Answer: \answerNA{}
    \item[] Justification:  The paper provides the compute resources used for the experiments, including the GPU type and implementation environment. The main experiments were conducted on L20X GPUs, and additional training details are described in the implementation details section.
    
\item {\bf Code of ethics}
    \item[] Question: Does the research conducted in the paper conform, in every respect, with the NeurIPS Code of Ethics \url{https://neurips.cc/public/EthicsGuidelines}?
    \item[] Answer: \answerYes{} 
    \item[] Justification: The research conforms to the NeurIPS Code of Ethics. The work uses standard public benchmark datasets and generated training samples only for model training and evaluation, without involving personally identifiable information or harmful deployment.

\item {\bf Broader impacts}
    \item[] Question: Does the paper discuss both potential positive societal impacts and negative societal impacts of the work performed?
    \item[] Answer: \answerYes{} 
    \item[] Justification: The paper discusses potential positive impacts, including improving few-shot object detection under limited annotation and cross-domain shifts. It also discusses practical limitations of the proposed method, such as the need for manually designed prompt templates and the potential dependence on generated samples for domain simulation.
    
\item {\bf Safeguards}
    \item[] Question: Does the paper describe safeguards that have been put in place for responsible release of data or models that have a high risk for misuse (e.g., pre-trained language models, image generators, or scraped datasets)?
    \item[] Answer: \answerNA{} 
    \item[] Justification: The paper does not release high-risk models, scraped datasets, or data that could directly enable misuse. The generated samples are used only for training and evaluation under standard benchmark settings.

\item {\bf Licenses for existing assets}
    \item[] Question: Are the creators or original owners of assets (e.g., code, data, models), used in the paper, properly credited and are the license and terms of use explicitly mentioned and properly respected?
    \item[] Answer: \answerYes{} 
    \item[] Justification: The paper cites the original sources of the public datasets, pretrained models, and codebases used in this work. We follow the official licenses and terms of use where available.

\item {\bf New assets}
    \item[] Question: Are new assets introduced in the paper well documented and is the documentation provided alongside the assets?
    \item[] Answer: \answerNA{} 
    \item[] Justification: The paper does not introduce new datasets, pretrained models, or benchmarks. Any released code is intended only to reproduce the proposed method on existing public datasets.

\item {\bf Crowdsourcing and research with human subjects}
    \item[] Question: For crowdsourcing experiments and research with human subjects, does the paper include the full text of instructions given to participants and screenshots, if applicable, as well as details about compensation (if any)? 
    \item[] Answer: \answerNA{} 
    \item[] Justification: The paper does not involve crowdsourcing experiments or research with human subjects. Therefore, there are no participant instructions, screenshots, or compensation details to report.

\item {\bf Institutional review board (IRB) approvals or equivalent for research with human subjects}
    \item[] Question: Does the paper describe potential risks incurred by study participants, whether such risks were disclosed to the subjects, and whether Institutional Review Board (IRB) approvals (or an equivalent approval/review based on the requirements of your country or institution) were obtained?
    \item[] Answer: \answerNA{} 
    \item[] Justification: The paper does not involve research with human subjects. Therefore, there are no participant risks, risk disclosures, or IRB approvals to report.

\item {\bf Declaration of LLM usage}
    \item[] Question: Does the paper describe the usage of LLMs if it is an important, original, or non-standard component of the core methods in this research? Note that if the LLM is used only for writing, editing, or formatting purposes and does \emph{not} impact the core methodology, scientific rigor, or originality of the research, declaration is not required.
    \item[] Answer: \answerYes{} 
    \item[] Justification: The paper describes the use of a VLM/multimodal generative model in the proposed prompt-driven domain simulation module. Its role in generating semantically consistent target-domain samples is explained in the method section.

\end{enumerate}

\end{document}